\documentclass[11pt,twocolumn]{article}

\usepackage[a4paper,margin=0.72in,top=0.64in,bottom=0.72in]{geometry}
\usepackage{times}
\usepackage{latexsym}
\usepackage[T1]{fontenc}
\usepackage[utf8]{inputenc}
\usepackage{microtype}
\usepackage{inconsolata}
\usepackage{graphicx}
\usepackage{booktabs}
\usepackage{amsmath}
\usepackage{amssymb}
\usepackage{caption}
\usepackage{enumitem}
\usepackage{placeins}
\usepackage{titlesec}
\usepackage[most]{tcolorbox}
\usepackage{flushend}
\usepackage{natbib}
\usepackage[colorlinks=true,citecolor=blue,linkcolor=blue,urlcolor=blue]{hyperref}

\setlist[itemize]{leftmargin=1.25em,topsep=1pt,itemsep=0pt,parsep=0pt}
\setlist[enumerate]{leftmargin=1.35em,topsep=1pt,itemsep=0pt,parsep=0pt}
\titlespacing*{\section}{0pt}{0.95em plus 0.15em minus 0.1em}{0.35em}
\titlespacing*{\subsection}{0pt}{0.7em plus 0.1em minus 0.08em}{0.25em}
\titlespacing*{\paragraph}{0pt}{0.55em plus 0.1em minus 0.08em}{0.6em}

\newcommand{\mvr}{\ensuremath{\mathrm{MVR}}}
\newcommand{\fs}{\ensuremath{\mathrm{FS}}}
\newcommand{\safe}{\ensuremath{c_{\mathrm{safe}}}}
\newcommand{\raised}{\ensuremath{c_{\mathrm{raised}}}}
\newcommand{\safei}{\ensuremath{c_{\mathrm{safe},i}}}
\newcommand{\raisedi}{\ensuremath{c_{\mathrm{raised},i}}}
\newcommand{\forcebench}{ForceBench}
\newcommand{\forcecheck}{ForceCheck}

\definecolor{appendixboxback}{RGB}{248,250,252}
\definecolor{appendixboxframe}{RGB}{174,181,192}
\newtcolorbox{appendixbox}[1]{%
  enhanced,
  breakable,
  colback=appendixboxback,
  colframe=appendixboxframe,
  colbacktitle=appendixboxback,
  coltitle=black,
  boxrule=0.35pt,
  arc=1pt,
  left=4pt,
  right=4pt,
  top=3pt,
  bottom=3pt,
  before skip=0.45em,
  after skip=0.55em,
  title={#1},
  fonttitle=\bfseries\small,
  fontupper=\footnotesize\raggedright,
}

\title{Relevant Is Not Warranted:\\[-0.15em]Evidence-Force Calibration for Cited RAG}
\author{
\textbf{Pin Qian$^{1,*}$ \quad
Su Wang$^{1,*}$ \quad
Xiaoyuan Wang$^{1,*}$ \quad
Yihang Chen$^{2,*}$ \quad
Wenxuan Xu$^{3}$}\\
\textbf{Qiaolin Yu$^{4}$ \quad
Shuhuai Lin$^{1}$ \quad
Sipeng Zhang$^{5}$ \quad
Junxian You$^{6}$ \quad
Xinpeng Wei$^{2}$}\\[0.5ex]
{\normalfont $^{1}$Carnegie Mellon University \quad
$^{2}$Georgia Institute of Technology \quad
$^{3}$Dartmouth College}\\
{\normalfont $^{4}$Cornell University \quad
$^{5}$University of California San Diego \quad
$^{6}$University of Glasgow}\\[0.5ex]
{\normalfont\small $^{*}$Equal contribution.}
}
\date{}

\newenvironment{frontabstract}{%
  \begin{center}\bfseries Abstract\end{center}
  \vspace{-0.45em}
  \begin{quote}\small\setlength{\parindent}{0pt}\setlength{\parskip}{0pt}
}{%
  \end{quote}
  \vspace{0.55em}
}

\makeatletter
\renewcommand{\maketitle}{%
  \begin{center}
    \vspace*{-0.18in}
    {\LARGE\bfseries \@title\par}
    \vspace{0.9em}
    {\large \@author\par}
  \end{center}
  \vspace{0.25em}
}
\makeatother

\begin{document}
\twocolumn[{
\maketitle

\begin{frontabstract}
Cited RAG evaluation often treats visible sources as a grounding signal, but a real, topically relevant citation can still under-warrant the attached wording. We study this diagnostic failure as \emph{citation laundering}: a related source is presented as warrant for an over-strong claim. We introduce \forcebench{}, a contrastive stress test for evidence-force calibration. Each item holds a cited passage fixed and pairs an evidence-calibrated claim with a localized force-raised variant across five operational axes: relation, modality, scope, temporal validity, and numeric specificity. A calibrated evaluator should score the evidence-calibrated claim higher. Headline experiments use a fixed, locality-filtered 198-pair evaluation set. A citation-presence sanity check is uninformative by design; token and entity overlap still violate monotonicity on 32.8--36.4\% of pairs. Across four reported model judges, standard generic support prompting is insufficient for this force-calibration stress test (aggregate MVR 47.2\%), while explicit warrant-strength prompting lowers MVR to 24.5\% but remains imperfect. We release the benchmark, prompts, outputs, and plug-in pipeline so citation evaluators can report monotonicity violation rate and force sensitivity alongside conventional support metrics.
\end{frontabstract}
}]

\begin{figure*}[t]
\centering
\captionsetup{width=0.9\textwidth}
\includegraphics[width=0.76\textwidth]{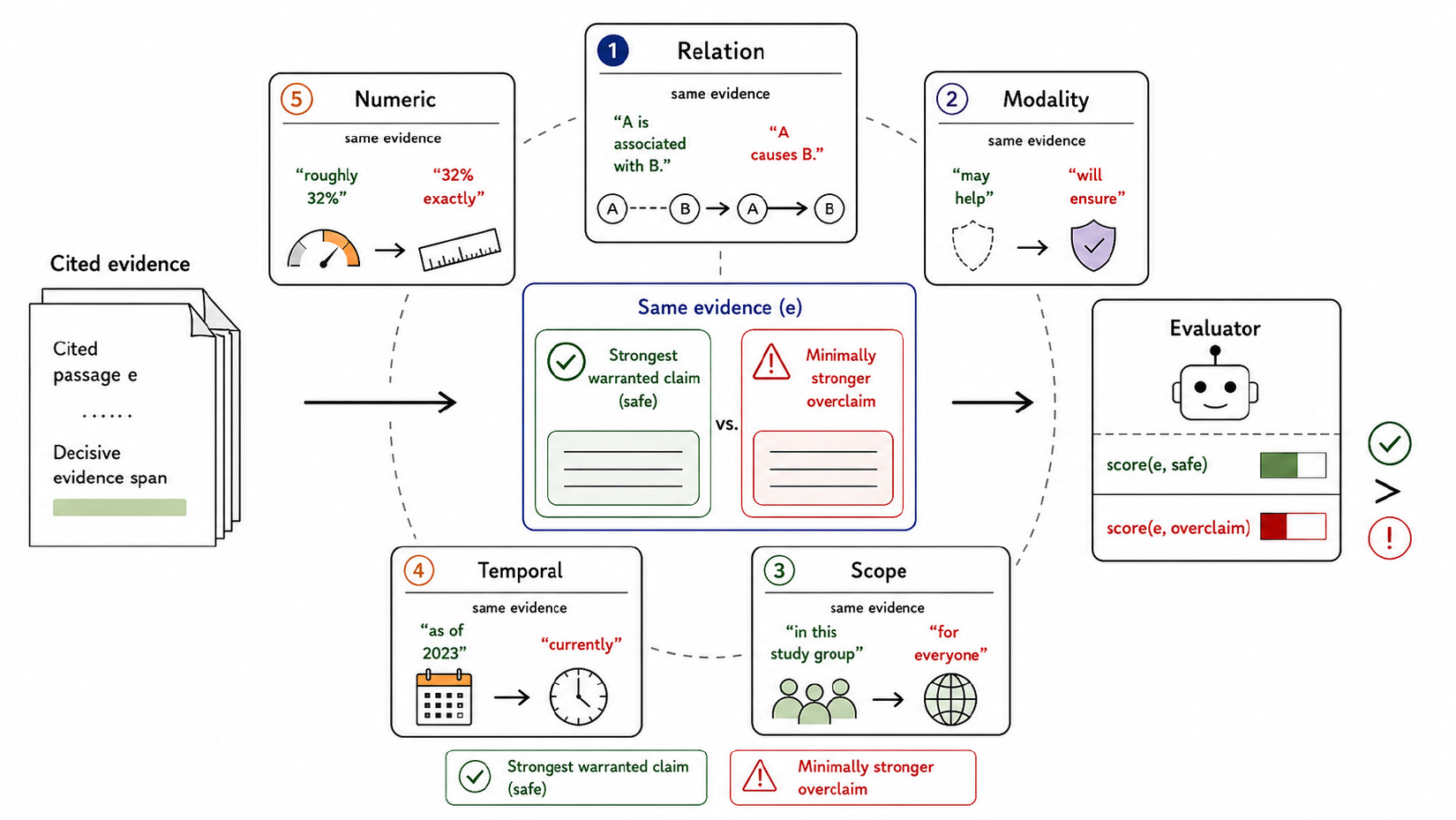}
\caption{\textbf{\forcebench{} evaluates citation laundering through contrastive monotonicity.}
The cited evidence is held fixed while the benchmark contrasts a strongest warranted claim with a minimally stronger overclaim. The same local test is instantiated across five force axes: relation, modality, scope, temporal validity, and numeric specificity. A calibrated evaluator should assign higher support to the warranted claim than to the overclaim.}
\label{fig:concept}
\end{figure*}

\section{Introduction}

Retrieval-augmented generation connects language generation to external evidence and provenance \citep{lewis-etal-2020-rag}. It has become a practical interface for domain-specific question answering over specialized document collections, including product-support settings \citep{sharma2024domainqa}, with recent examples also studying financial-report question answering and retrieval \citep{cheng2026resolvingrobustnessprecisiontradeofffinancial,cheng2026enhancingfinancialreportquestionanswering}. In cited RAG, citations have become a visible grounding interface: browser-assisted QA, quote-supported QA, and citation-augmented generation systems attach references or supporting evidence to generated answers, inviting users to treat cited claims as grounded \citep{nakano-etal-2021-webgpt,menick-etal-2022-verified,gao-etal-2023-enabling,rashkin-etal-2023-measuring}. This presentation can compress a hard verification problem into a visible cue: this sentence has a source. The cue is useful, but incomplete. A citation can be present, relevant, and still too weak, narrow, stale, or imprecise for the sentence it is asked to support.

Consider an answer that says a drug exposure \emph{causes} an adverse event while the cited source reports only an observational association and explicitly notes that causality could not be established. Association-to-causation shifts are a known form of scientific overstatement \citep{sumner-etal-2014-exaggeration}; in high-stakes health settings, the distinction licenses different downstream inferences. The citation is real. The entities match. A citation-presence metric passes it, and many retrieval or overlap checks see the right topic. The evidence licenses a weaker claim: association, not causation. The generated answer has increased the \emph{force} of the cited evidence.

The same pattern appears outside medical causality. In our adjudicated examples, one cited passage says that men \emph{may} have been more susceptible to a historical plague; the answer cites it to state that men \emph{were} more susceptible. A public-health passage says that Ebola's $R_0$ typically sits around 1.5 to 2.0; the answer cites it for the point claim that Ebola's $R_0$ is 2.0. A clinical passage reports a trial in children with convulsive status epilepticus; the answer cites it for a claim about status epilepticus patients generally. In each case, the citation is related enough to appear plausible, but the claim asks it to carry too much evidential force.

We call this failure \textbf{citation laundering}, using the term narrowly for cases where a related citation makes an over-strong claim appear warranted. The citation exists, is recoverable, and is sufficiently related to pass surface inspection, yet it is too weak for the claim's relation, modality, scope, temporal status, or numeric precision. The failure mode is that surface relevance can mask insufficient warrant.

\paragraph{Relation to hallucination.}
Citation laundering is a factuality failure in the claim-evidence relation. It differs from factuality and hallucination benchmarks that evaluate false, unsupported, or unverifiable model outputs more broadly \citep{lin-etal-2022-truthfulqa,li-etal-2023-halueval}: here the source exists, is recoverable, and is topically related, while the displayed citation under-warrants the force of the claim. A detector that only checks whether each claim has some cited source will pass every such claim by construction. We therefore frame citation laundering at the claim-citation pair level.

This paper introduces \textbf{evidence-force calibration} for cited RAG. Given cited evidence $e$ and claim $c$, we ask whether the force of $c$ is no greater than the force licensed by $e$. We operationalize this with a contrastive monotonicity test (Figure~\ref{fig:concept}): if the same evidence supports an evidence-calibrated claim $\safe$, then a localized force-raised variant $\raised$ should receive lower support:
\begin{equation}
    score(e,\safe) > score(e,\raised).
\end{equation}
If a judge scores ``may help'' and ``will help'' equally against evidence that only says ``may,'' it loses the evidence-force ordering even if it recognizes the topic.

This makes the evaluation target operational. A useful cited-generation system should preserve citation-warranted wording such as ``may help'' while rejecting stronger wording such as ``will ensure,'' because ranking, filtering, and repair need calibrated claims alongside unsupported-claim warnings.

Our contributions are:
\begin{enumerate}
    \item We formalize citation laundering as a mismatch between claim force and evidence force, distinct from missing citations, irrelevant retrieval, and ordinary unsupported generation.
    \item We introduce \forcebench{}, a diagnostic contrastive benchmark of local force contrasts across relation, modality, scope, temporal, and numeric axes.
    \item We define monotonicity violation rate and force sensitivity for evaluating whether citation evaluators respect local increases in claim force.
    \item We construct an adjudicated benchmark and report headline results on a fixed locality-filtered evaluation set.
    \item We evaluate deterministic baselines and four deployed model judges, finding that standard generic citation-support prompting is insufficient for this force-calibration stress test; explicit warrant-strength prompting helps but remains imperfect.
\end{enumerate}

\section{Evidence-Force Calibration}

Attribution and citation evaluation typically ask whether cited evidence supports a claim or answer \citep{rashkin-etal-2023-measuring,gao-etal-2023-enabling,yue-etal-2023-automatic,citeeval2025}. \forcebench{} asks a stricter question: does the evidence warrant the claim at the same level of relation, modality, scope, temporal validity, and numeric specificity?

We model claim force as a structured categorical tuple:
\begin{equation}
\begin{aligned}
claim\_force(c) = \{&\text{relation}, \text{modality}, \text{scope},\\
&\text{temporal}, \text{numeric}\}.
\end{aligned}
\end{equation}
Evidence force is the corresponding warrant licensed by the cited passage, including stated limitations and context. A claim is force-calibrated if:
\begin{equation}
claim\_force(c) \leq evidence\_force(e),
\end{equation}
where $\leq$ is an evidence-relative partial order: the claim does not exceed the cited passage on any included axis. Many claims are incomparable. \forcebench{} only constructs pairs that preserve the same core proposition and differ by a localized force increase on one primary axis.

\paragraph{Axes.}
The five axes are operational and intentionally non-exhaustive. We select dimensions that satisfy a practical test: holding the cited evidence fixed, a claim can remain topically related while exceeding what the citation warrants through a local increase in relation strength, certainty, scope, temporal status, or numeric specificity. The axes draw on adjacent evaluation lines without treating any one taxonomy as exhaustive. Attribution and citation evaluation motivate the local support question \citep{rashkin-etal-2023-measuring,citeeval2025}. Atomic and long-form factuality motivate claim-level granularity \citep{min-etal-2023-factscore,wei-etal-2024-longform}. Hedge and uncertainty work informs the modality axis \citep{farkas-etal-2010-conll,yona-etal-2024-large}. Temporal annotation and event factuality motivate temporal contrasts \citep{sauri-pustejovsky-2009-factbank,pustejovsky-etal-2003-timeml}; fact-verification benchmarks motivate evidential-status contrasts \citep{thorne-etal-2018-fever,aly-etal-2021-feverous}. Scientific overstatement work motivates relation-strength contrasts such as association-to-causation \citep{sumner-etal-2014-exaggeration}. We exclude dimensions such as source authority, source freshness as a property of the citation itself, citation placement, and aggregation across multiple sources because they require answer-level or source-set judgments beyond a local single-citation force test. \forcebench{} makes the included axes citation-specific by asking whether the displayed source licenses the stronger wording.

\forcebench{} focuses on five axes:
\begin{itemize}
    \item \textbf{Relation:} association, mention, or risk becomes causation, prevention, proof, or obligation.
    \item \textbf{Modality:} possible, conditional, preliminary, or suggestive evidence becomes definite, necessary, or guaranteed.
    \item \textbf{Scope:} a claim licensed for a subgroup, jurisdiction, study population, version, or institution becomes general.
    \item \textbf{Temporal:} dated, predicted, or as-of evidence becomes a current or timeless claim.
    \item \textbf{Numeric:} approximate, ranged, or bounded evidence becomes an exact endpoint or point estimate.
\end{itemize}

\paragraph{Boundary.}
Citation laundering covers the subset of citation errors where a related source under-warrants a localized force increase. Fabricated quotations, wrong-entity claims, missing links, and broad unsupported additions are important failures with different diagnostic structure. \forcebench{} isolates the local property that support should decrease when claim force increases while the cited evidence remains fixed.

\section{\forcebench{}}

\forcebench{} uses one cited evidence passage and one atomic claim as its unit. Each item contains the cited passage, a decisive evidence span, an evidence-calibrated claim, a localized force-raised claim, a primary force axis, severity, and a repair. The calibrated claim is wording that annotators judge to be licensed by the cited passage. The force-raised claim preserves the same local proposition but exceeds the evidence on one primary axis.

\subsection{Construction}

\forcebench{} is built through a single local-warrant annotation protocol. We start from cited claim-evidence contexts sampled from existing cited QA and attribution resources. One source is AttributionBench-derived rows \citep{attributionbench2024}, spanning ExpertQA \citep{malaviya-etal-2024-expertqa}, LFQA-style long-form QA \citep{chen-etal-2024-understanding}, and Stanford-GenSearch-style verifiability settings \citep{liu-etal-2023-evaluating}. We also include contexts from AttributedQA \citep{bohnet-etal-2023-attributedqa}, GaRAGe \citep{sorodoc-etal-2025-garage}, and additional ExpertQA examples. The benchmark is a stress test, with sampling targeted at contexts whose cited passages expose local boundaries in relation, modality, scope, temporal status, or numeric specificity.

For each selected context, we draft a candidate row containing the cited passage, a decisive evidence span, an evidence-calibrated claim, a localized force-raised claim, a repair, a rationale, and proposed axis and severity metadata. These proposed fields are pre-annotation suggestions. Every candidate row is independently annotated by two research assistants under the same local-warrant guideline. The two independent reviews are then reconciled through adjudication. Rows that pass adjudication undergo quality control of the evidence span, calibrated claim, force-raised claim, repair, primary axis, and severity; only rows passing both adjudication and quality control are eligible for inclusion.

The protocol asks annotators to answer one local question: what wording does this displayed citation license? Annotators identify the decisive evidence span, verify an evidence-calibrated claim, test a stronger variant against the same passage, assign one primary force axis when the contrast is local, and write a repair that restores the citation-warranted wording. The central guideline is to ignore external truth: a claim may be true in the world and still be a \forcebench{} force gap if the displayed citation does not warrant it.

We use \emph{localized} as an operational criterion rather than a formal edit-distance guarantee. A pair is admissible only if the two claims preserve the same cited passage, answer context, main entity, event, and topical relation, while differing primarily in one force dimension. Candidates are rejected if the calibrated claim is not supported, the stronger claim is already supported, the contrast changes entities or events, combines multiple force shifts, introduces an unrelated unsupported fact, requires outside knowledge, or lacks a clear evidence span.

\subsection{Quality Control}

Force gaps are close to their positive counterparts, so we use two independent annotations for every candidate row, followed by adjudication, evidence-span checks, locality checks, and provenance/duplicate checks. On pre-adjudication row-level decisions, the two research assistants achieve Cohen's $\kappa=0.78$. The audit asks whether the cited evidence warrants the calibrated formulation while under-warranting the stronger formulation.

Final promotion is deliberately conservative. A row enters the benchmark only if the evidence span is decisive, the calibrated claim is supported, the stronger claim is under-warranted, and the contrast is a local force increase with no ordinary unsupported addition. Because the candidate pool is built to cover the five force axes, surplus rows from overrepresented axes are capped before benchmark reporting to avoid treating selected counts as prevalence evidence. A final locality audit removed two accepted rows whose force-raised claims introduced ordinary unsupported additions beyond the intended force shift. The retained benchmark therefore contains 229 adjudicated pairs with per-axis counts between 39 and 51 (Table~\ref{tab:data_stats}); all headline results use a fixed 198-pair locality-filtered evaluation set. When forming retained and evaluation sets, we stratify by force axis and source provenance and use a stable hash for tie-breaking, separating candidate collection from the metric denominator. Appendix~\ref{sec:appendix-data} gives the candidate accounting, split/filter rule, and annotation statistics; Appendix~\ref{sec:appendix-annotation} gives the human annotation guidelines.

\begin{table}[t]
\centering
\small
\begin{tabular}{lrrrrr}
\toprule
Axis & Retained & Eval & S1 & S2 & S3 \\
\midrule
Relation & 48 & 40 & 0 & 38 & 2 \\
Modality & 48 & 39 & 5 & 33 & 1 \\
Scope & 51 & 40 & 1 & 34 & 5 \\
Temporal & 39 & 39 & 10 & 27 & 2 \\
Numeric & 43 & 40 & 16 & 23 & 1 \\
\midrule
Total & 229 & 198 & 32 & 155 & 11 \\
\bottomrule
\end{tabular}
\caption{\textbf{\forcebench{} data summary.}
Retained counts are adjudicated pairs kept for the benchmark pool after capping 52 surplus relation candidates and removing two nonlocal rows; Eval is the fixed locality-filtered evaluation set. S1--S3 are severity counts in Eval.}
\label{tab:data_stats}
\end{table}

\section{Metrics}

For an evaluator that assigns support scores, a monotonicity violation occurs when the force-raised claim receives support greater than or equal to the evidence-calibrated claim. We define monotonicity violation rate:
\begin{equation}
\mvr = \frac{1}{N}\sum_i
\mathbf{1}[s_i^{\mathrm{raised}}\geq s_i^{\mathrm{safe}}].
\end{equation}
Here $s_i^{\mathrm{safe}}=score(e_i,\safei)$ and $s_i^{\mathrm{raised}}=score(e_i,\raisedi)$.
Lower is better. We also report force sensitivity:
\begin{equation}
\fs = \frac{1}{N}\sum_i
(s_i^{\mathrm{safe}}-s_i^{\mathrm{raised}}).
\end{equation}
Higher is better. \mvr{} captures whether the evaluator preserves strict ordering; \fs{} captures the average margin. The strict inequality is deliberate: a tie between the evidence-calibrated claim and the force-raised variant gives no preference to the warranted wording for ranking, filtering, or repair. Both metrics are therefore necessary: a model may increase margins while still assigning both claims the same discrete label.

Categorical judges are mapped to scores. For generic support prompts, \texttt{fully\_supported} receives 1, \texttt{partially\_supported} receives 0.5, and unsupported, contradicted, or irrelevant labels receive 0. For force-aware prompts, \texttt{force\_calibrated} receives 1, while force-gap, contradicted, or irrelevant labels receive 0.
If a model returns an unparsable response or refuses after retry for either member of a pair, we count that pair as a monotonicity violation with margin 0. This conservative policy keeps denominators fixed and treats evaluator non-compliance as a failure to provide an actionable preference.

\begin{figure}[tb]
\centering
\includegraphics[width=\linewidth]{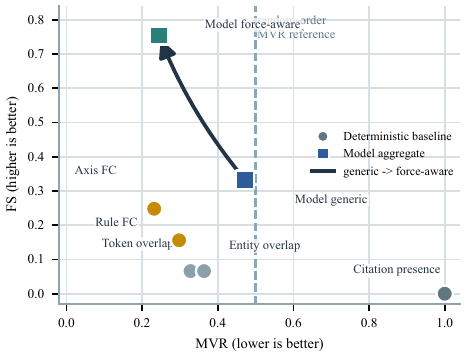}
\caption{\textbf{Main MVR--FS tradeoff on the 198-pair evaluation set.}
Each point is an evaluator. Lower MVR and higher FS are better, so the desired region is upper-left. Circles show deterministic baselines; squares show aggregate model judges. The arrow shows generic-to-force-aware prompting, which moves the model aggregate toward fewer monotonicity violations and larger force margins.}
\label{fig:main_results}
\end{figure}

\section{Experiments}

We evaluate whether common citation signals and strong LLM judges preserve force monotonicity.

\subsection{Baselines}

We report citation presence, token overlap, topical entity overlap, a rule-based \forcecheck{}, an axis-aware lexical \forcecheck{} prototype, and four model judges. The axis-aware prototype starts from token overlap and applies penalties when the claim removes limiters visible in the evidence, such as changing ``may'' to ``will,'' a range to a point estimate, or a dated statement to a current claim. It is an interpretable diagnostic baseline for testing whether visible force markers matter. Its limitations include paraphrase, implicit scope, and context-dependent evidence boundaries.

\paragraph{Model judge selection.}
We evaluate four high-capability LLM judges: Claude Sonnet 4.6, GPT-5.5, GLM 5.1, and Qwen 3.6 Flash. The panel is intentionally limited. It targets capable deployed judges and asks whether they preserve calibrated-over-force-raised ordering under fixed evidence when prompted with a standard support question. Using multiple model families gives a limited cross-interface check of the observed pattern.

We evaluated GPT-5.5, Claude Sonnet 4.6, GLM 5.1, and Qwen 3.6 Flash through API calls recorded in the run manifests. Each JSON-only prompt is applied to both the evidence-calibrated claim and the force-raised claim for every pair on the same 198-pair locality-filtered evaluation set. Each item is submitted as a fresh deterministic, stateless request with tool access disabled and a 1024-token output cap. Full model identifiers, API settings, run dates, parser settings, and retry policy are provided in Appendix~\ref{app:model_config}.

This evaluator suite targets the paper's narrow local-ordering claim under fixed evidence. \mvr{} and \fs{} are plug-in stress-test metrics: any attribution or citation evaluator can be run on the same contrastive pairs. The reported suite spans the main signals needed for this diagnostic: citation appearance, topical overlap, visible force-shift heuristics, generic support judging, and explicit force-aware judging. Broader learned or task-specific citation evaluators remain important future plug-ins.

\subsection{Main Results}

Figure~\ref{fig:main_results} plots monotonicity violation against force sensitivity. Citation presence is a sanity check: it never distinguishes the paired claims and therefore has 100\% MVR by construction. Token and entity overlap test whether topical similarity is enough to preserve monotonicity; both remain low-sensitivity despite using more information than citation presence. The axis-aware lexical prototype is the strongest deterministic baseline (MVR 0.232), but it is hand-aligned with visible force markers and still leaves substantial failures, especially when force is implicit.

Table~\ref{tab:main_results} summarizes deterministic baselines and aggregate model-judge results for the two reported model prompts. The pooled model-judge result is the central stress test. With a generic support prompt, the four-judge panel has MVR 0.472 (374/792), and force sensitivity is 0.333. Under an explicitly force-aware prompt, MVR drops to 0.245 (194/792), and force sensitivity increases to 0.754. This paired comparison should be read as a prompt-conditioned diagnostic, because the generic and force-aware prompts ask different questions. It shows that ordinary support wording is insufficient for this force-calibration stress test, and that making warrant strength explicit improves ordering on the same model-pair units.

\begin{table}[tb]
\centering
\small
\setlength{\tabcolsep}{4pt}
\begin{tabular}{llrr}
\toprule
Evaluator & Prompt & MVR & FS \\
\midrule
\multicolumn{4}{l}{\textit{Deterministic baselines}} \\
Citation presence & -- & 1.000 & 0.000 \\
Token overlap & -- & 0.328 & 0.066 \\
Entity overlap & -- & 0.364 & 0.066 \\
Rule \forcecheck{} & -- & 0.298 & 0.156 \\
Axis \forcecheck{} & -- & 0.232 & 0.248 \\
\midrule
\multicolumn{4}{l}{\textit{Reported model judges}} \\
Model aggregate & Generic support & 0.472 & 0.333 \\
Model aggregate & Force-aware & 0.245 & 0.754 \\
\bottomrule
\end{tabular}
\caption{\textbf{Main MVR and FS on the 198-pair evaluation set.}
Lower MVR and higher FS are better. Model rows report aggregate results across the four reported judges. Parse/refusal pairs are counted as monotonicity violations with zero margin.}
\label{tab:main_results}
\end{table}

\subsection{Prompt-Rubric Ablation}
\label{sec:prompt_ablation}

\begin{figure}[tb]
\centering
\includegraphics[width=\linewidth]{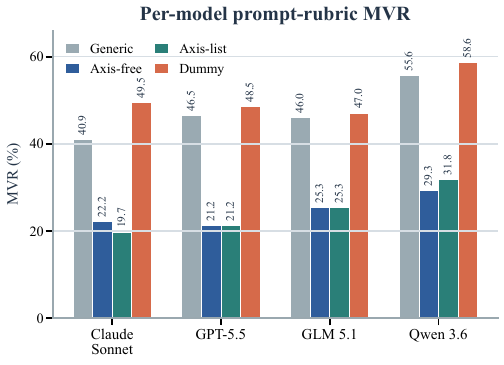}
\caption{\textbf{Per-model prompt-rubric MVR.}
Each model has four bars: generic support, axis-free force, axis-list force-aware, and dummy-axis control. Lower MVR is better.}
\label{fig:per_model}
\end{figure}

The reported force-aware prompt names the same five force axes used to construct \forcebench{}, so we test whether axis names drive the result. Table~\ref{tab:prompt_ablation_main} compares two controls: a minimal axis-free force prompt that keeps the warrant-strength instruction while omitting the benchmark taxonomy, and a dummy-axis control that uses irrelevant rubric dimensions. The axis-free prompt reduces aggregate MVR from 47.2\% to 24.5\%, while the axis-list prompt also yields 24.5\% MVR and raises FS to 0.754. The dummy-axis control rises to 50.9\%. The prompt gain therefore tracks explicit warrant-strength framing. We interpret it as prompt-conditioned task framing for this stress test.

\begin{table}[tb]
\centering
\small
\setlength{\tabcolsep}{4pt}
\begin{tabular}{lrr}
\toprule
Prompt & MVR (\%) & FS \\
\midrule
Generic support & 47.2 & 0.333 \\
Minimal axis-free force & 24.5 & 0.753 \\
Axis-list force-aware & 24.5 & 0.754 \\
Dummy-axis control & 50.9 & 0.480 \\
\bottomrule
\end{tabular}
\caption{\textbf{Aggregate prompt-rubric ablation.}
Lower MVR and higher FS are better. The axis-free prompt omits the benchmark axis taxonomy; the dummy-axis control uses irrelevant rubric dimensions.}
\label{tab:prompt_ablation_main}
\end{table}

Figure~\ref{fig:per_model} shows the per-model prompt-rubric pattern. Both force prompts improve over generic support prompting for all four judges, while the dummy-axis control often worsens MVR. Table~\ref{tab:prompt_ablation_main} reports the corresponding aggregate MVR and FS values. We interpret this as a prompt-conditioned result for this four-judge panel.

\subsection{Axis Behavior}

\begin{figure}[tb]
\centering
\includegraphics[width=\linewidth]{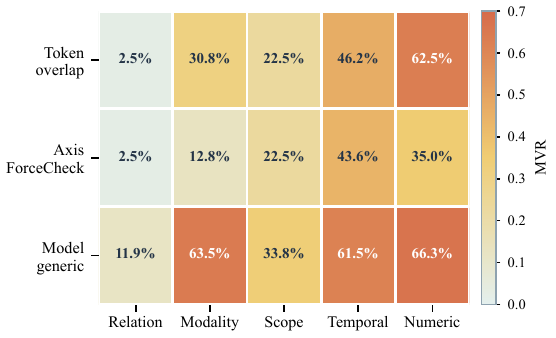}
\caption{\textbf{Axis-level MVR heatmap on the 198-pair evaluation set.}
Darker and warmer cells indicate higher violation rates. Deterministic rows use 39--40 pairs per axis; the aggregate generic-support model row uses 156--160 model-pair units per axis. Relation shifts are easiest because they are often lexical, while modality, numeric, and temporal shifts are harder under ordinary support prompting.}
\label{fig:axis_radar}
\end{figure}

Axis behavior is reported in Figure~\ref{fig:axis_radar}. Relation shifts are comparatively visible because an association-to-causation change is often lexical; the aggregate generic-support row has its lowest MVR on relation. Numeric, temporal, and modality shifts are harder: they often require attending to ranges, dates, hedges, or stated uncertainty rather than matching entities. The heatmap makes this asymmetry visible across deterministic baselines and ordinary support prompting.

\section{Discussion}

\paragraph{Warrant calibration.}
The central lesson concerns warrant calibration. A citation evaluator that rejects both variants may be safer than one that accepts both, yet it still cannot recover citation-warranted wording. The force-aware prompt improves ordering overall while still leaving residual violations. Evidence-force calibration asks for a precise capability: keep the warranted claim and reject only the extra force.

\paragraph{Residual prompt-conditioned errors.}
In this fixed evaluation set, with these prompts and single-run model calls, the aggregate force-aware MVR remains 0.245. Closing those residual errors likely requires evaluator-side supervision, architectures, or training data targeted at warrant calibration, beyond prompt engineering alone.

\paragraph{Limits of the diagnostic heuristic.}
The axis-aware \forcecheck{} result tests whether visible limiters are informative. The heuristic is hand-aligned with the benchmark axes and depends on explicit lexical cues; it performs well on relation and modality cases with overt markers, and still violates monotonicity on 35.0\% of numeric and 43.6\% of temporal examples. This residual pattern is important: many force gaps involve a relationship between the wording of the claim and the limitations of the cited passage.

\paragraph{Why contrastive pairs matter.}
Absolute support labels are noisy because annotators and models differ in how much context they require. The contrastive design holds evidence fixed and asks only for an ordering. Observed cited-output examples motivate the failure pattern, while source-grounded local contrasts remove retrieval quality, answer style, and external truth as explanations. This makes the scientific claim crisp: a citation evaluator should give lower support to a stronger claim when the extra force exceeds what the source licenses.

\section{Related Work}

\forcebench{} is complementary to attribution and citation evaluation. AIS defines attribution as verification against identified sources \citep{rashkin-etal-2023-measuring}; ALCE evaluates citation-augmented long-form QA along correctness and citation quality dimensions \citep{gao-etal-2023-enabling}; and automatic attribution evaluators such as AttrScore investigate LLM and finetuned support judgments for cited claims \citep{yue-etal-2023-automatic}. AttributionBench-style data \citep{attributionbench2024}, broad RAG diagnostics such as RAGChecker \citep{ragchecker2024}, and citation-specific evaluators such as CiteEval \citep{citeeval2025} provide broader citation-quality testbeds. CiteEval is especially close in motivation because it makes citation assessment more fine-grained; \forcebench{} fixes the cited passage and measures a local monotonicity property under a controlled force increase. Citation-context and citation-intent work studies how scientific papers use prior work \citep{cohan-etal-2019-structural}; \forcebench{} asks whether a displayed citation warrants the force of a generated claim.
Adjacent robustness work studies reward-model behavior under perturbations and attention-based failure modes \citep{zang2025reward,zang2025alleviating}, as well as safeguard robustness against adversarial prompts \citep{lin2026reflect}.

Claim-level factuality and hallucination benchmarks ask whether generated content is truthful, hallucinated, or factually precise, including TruthfulQA, HaluEval, FActScore, and SAFE/Long-form factuality \citep{lin-etal-2022-truthfulqa,li-etal-2023-halueval,min-etal-2023-factscore,wei-etal-2024-longform}. NLI and fact-verification datasets formalize support as entailment, contradiction, or neutrality \citep{bowman-etal-2015-large,thorne-etal-2018-fever,aly-etal-2021-feverous}, and hedge/uncertainty work motivates force distinctions \citep{farkas-etal-2010-conll,yona-etal-2024-large}. These lines target absolute truth, support classification, or uncertainty expression. \forcebench{} conditions on a real, recoverable, topically related citation and asks whether support is monotonic when only claim force increases. Appendix~\ref{sec:appendix-positioning} summarizes this boundary with adjacent evaluation lines.

The closest conceptual line is scientific overstatement, especially health-science exaggeration analysis \citep{sumner-etal-2014-exaggeration}. \forcebench{} transfers that concern to generated cited answers, where an inline citation may make a claim appear warranted. The unit is the claim-citation pair in RAG answers. The label target is whether the displayed citation licenses the stronger wording. The metric is therefore contrastive monotonicity under fixed cited evidence.

\section{Conclusion}

\forcebench{} targets a missing layer in cited RAG evaluation: evidence warrant. A citation can be real and relevant while still failing to license the force of the attached claim. By holding evidence fixed and applying a localized force increase, \forcebench{} turns this failure into a direct monotonicity test. On the fixed locality-filtered 198-pair evaluation set, overlap baselines violate monotonicity on 32.8--36.4\% of pairs, and standard generic-support prompting yields aggregate model-judge MVR 47.2\%. Explicit warrant-strength prompting reduces aggregate model MVR to 24.5\% on the same model-pair units, but residual failures remain. For practitioners and benchmark users, the implication is diagnostic: citation-presence checks should be treated as a minimal sanity check and supplemented with warrant-sensitive verification on stress tests such as \forcebench{}. For evaluators, \mvr{} and \fs{} provide plug-in stress-test metrics alongside conventional support accuracy. For researchers, the remaining errors point toward evaluator-side training and supervision for warrant calibration.

\section*{Limitations}

\forcebench{} is a compact diagnostic benchmark. Its MVR numbers are stress-test error rates under fixed evidence, with prevalence estimation outside scope. The examples are selected for visible local warrant boundaries, so prompt gains reflect this targeted evaluation set. The force-aware and axis-free prompts explicitly define the warrant-strength target; their gains show the effect of prompt-conditioned task framing in this stress test. Larger releases should broaden domain, source, and language coverage. Future work should also extend evidence-force calibration beyond text-only claim-passage pairs to multimodal cited outputs, where citation warrant may depend jointly on visual evidence, textual claims, and domain-specific interpretation; cross-domain VQA adaptation work such as CATCH illustrates practical domain-shift challenges in medical, remote-sensing, chart, and math-diagram settings \citep{li2025catch}. Locality is an operational annotation judgment rather than a formal edit-distance guarantee; a final audit removed two nonlocal rows, and the release reports those IDs and reasons, but borderline cases may remain. The evaluator suite covers citation presence, lexical/entity overlap, rule/axis \forcecheck{} prototypes, and four model judges; CiteEval-style, AutoAIS-style, and learned attribution baselines remain useful broader coverage \citep{citeeval2025,gao-etal-2023-enabling,yue-etal-2023-automatic}. Model-judge results are single-run measurements of closed or hosted systems under the prompts reported here; model drift, decoding changes, and prompt-target alignment can change absolute and relative MVR values. Tone and politeness prompt effects can also vary across model families and task domains \citep{cai2025tone}. Repair fields exist for accepted items, while repair success is outside the main evaluation. \forcebench{} isolates single English claim-citation pairs; coverage excludes multi-hop synthesis, aggregation across multiple sources, source authority, citation recency as a source property, citation placement, and answer-level usefulness.

\section*{Ethical Considerations}

\forcebench{} is an evaluation resource for identifying cases where cited evidence under-warrants the force of a generated claim. Use of the benchmark should stay within citation-warrant evaluation, with population prevalence outside scope. Some examples involve medical, legal, or policy-sensitive content; labels concern citation warrant rather than real-world truth or advice. Released rows preserve source provenance and use citation snippets for research evaluation; users should respect the terms of the underlying source collections when redistributing passages. The research assistants judged only local citation warrant and were instructed to avoid medical, legal, or policy advice. Data release should preserve annotation decisions, rejected-boundary labels, and the distinction between final adjudicated labels and diagnostic artifacts.

\bibliographystyle{acl_natbib}
\bibliography{forcebench_references}

@article{rashkin-etal-2023-measuring,
  title = {Measuring Attribution in Natural Language Generation Models},
  author = {Rashkin, Hannah and Nikolaev, Vitaly and Lamm, Matthew and Aroyo, Lora and Collins, Michael and Das, Dipanjan and Petrov, Slav and Tomar, Gaurav Singh and Turc, Iulia and Reitter, David},
  journal = {Computational Linguistics},
  volume = {49},
  number = {4},
  pages = {777--840},
  address = {Cambridge, MA},
  publisher = {MIT Press},
  year = {2023},
  doi = {10.1162/coli_a_00486},
  url = {https://aclanthology.org/2023.cl-4.2/}
}

@inproceedings{lewis-etal-2020-rag,
  title = {Retrieval-Augmented Generation for Knowledge-Intensive {NLP} Tasks},
  author = {Lewis, Patrick and Perez, Ethan and Piktus, Aleksandra and Petroni, Fabio and Karpukhin, Vladimir and Goyal, Naman and K{\"u}ttler, Heinrich and Lewis, Mike and Yih, Wen-tau and Rockt{\"a}schel, Tim and Riedel, Sebastian and Kiela, Douwe},
  booktitle = {Advances in Neural Information Processing Systems 33},
  pages = {9459--9474},
  year = {2020},
  url = {https://proceedings.neurips.cc/paper/2020/hash/6b493230205f780e1bc26945df7481e5-Abstract.html}
}

@article{nakano-etal-2021-webgpt,
  title = {{WebGPT}: Browser-assisted Question-answering with Human Feedback},
  author = {Nakano, Reiichiro and Hilton, Jacob and Balaji, Suchir and Wu, Jeff and Ouyang, Long and Kim, Christina and Hesse, Christopher and Jain, Shantanu and Kosaraju, Vineet and Saunders, William and Jiang, Xu and Cobbe, Karl and Eloundou, Tyna and Krueger, Gretchen and Button, Kevin and Knight, Matthew and Chess, Benjamin and Schulman, John},
  journal = {arXiv preprint arXiv:2112.09332},
  year = {2021},
  doi = {10.48550/arXiv.2112.09332},
  url = {https://arxiv.org/abs/2112.09332}
}

@article{menick-etal-2022-verified,
  title = {Teaching Language Models to Support Answers with Verified Quotes},
  author = {Menick, Jacob and Trebacz, Maja and Mikulik, Vladimir and Aslanides, John and Song, Francis and Chadwick, Martin and Glaese, Mia and Young, Susannah and Campbell-Gillingham, Lucy and Irving, Geoffrey and McAleese, Nat},
  journal = {arXiv preprint arXiv:2203.11147},
  year = {2022},
  doi = {10.48550/arXiv.2203.11147},
  url = {https://arxiv.org/abs/2203.11147}
}

@inproceedings{gao-etal-2023-enabling,
  title = {Enabling Large Language Models to Generate Text with Citations},
  author = {Gao, Tianyu and Yen, Howard and Yu, Jiatong and Chen, Danqi},
  booktitle = {Proceedings of the 2023 Conference on Empirical Methods in Natural Language Processing},
  pages = {6465--6488},
  address = {Singapore},
  publisher = {Association for Computational Linguistics},
  year = {2023},
  doi = {10.18653/v1/2023.emnlp-main.398},
  url = {https://aclanthology.org/2023.emnlp-main.398/}
}

@inproceedings{yue-etal-2023-automatic,
  title = {Automatic Evaluation of Attribution by Large Language Models},
  author = {Yue, Xiang and Wang, Boshi and Chen, Ziru and Zhang, Kai and Su, Yu and Sun, Huan},
  booktitle = {Findings of the Association for Computational Linguistics: EMNLP 2023},
  pages = {4615--4635},
  address = {Singapore},
  publisher = {Association for Computational Linguistics},
  year = {2023},
  doi = {10.18653/v1/2023.findings-emnlp.307},
  url = {https://aclanthology.org/2023.findings-emnlp.307/}
}

@inproceedings{lin-etal-2022-truthfulqa,
  title = {{TruthfulQA}: Measuring How Models Mimic Human Falsehoods},
  author = {Lin, Stephanie and Hilton, Jacob and Evans, Owain},
  booktitle = {Proceedings of the 60th Annual Meeting of the Association for Computational Linguistics (Volume 1: Long Papers)},
  pages = {3214--3252},
  address = {Dublin, Ireland},
  publisher = {Association for Computational Linguistics},
  year = {2022},
  doi = {10.18653/v1/2022.acl-long.229},
  url = {https://aclanthology.org/2022.acl-long.229/}
}

@inproceedings{li-etal-2023-halueval,
  title = {{HaluEval}: A Large-Scale Hallucination Evaluation Benchmark for Large Language Models},
  author = {Li, Junyi and Cheng, Xiaoxue and Zhao, Xin and Nie, Jian-Yun and Wen, Ji-Rong},
  booktitle = {Proceedings of the 2023 Conference on Empirical Methods in Natural Language Processing},
  pages = {6449--6464},
  address = {Singapore},
  publisher = {Association for Computational Linguistics},
  year = {2023},
  doi = {10.18653/v1/2023.emnlp-main.397},
  url = {https://aclanthology.org/2023.emnlp-main.397/}
}

@inproceedings{min-etal-2023-factscore,
  title = {{FA}ct{S}core: Fine-grained Atomic Evaluation of Factual Precision in Long Form Text Generation},
  author = {Min, Sewon and Krishna, Kalpesh and Lyu, Xinxi and Lewis, Mike and Yih, Wen-tau and Koh, Pang and Iyyer, Mohit and Zettlemoyer, Luke and Hajishirzi, Hannaneh},
  booktitle = {Proceedings of the 2023 Conference on Empirical Methods in Natural Language Processing},
  pages = {12076--12100},
  address = {Singapore},
  publisher = {Association for Computational Linguistics},
  year = {2023},
  doi = {10.18653/v1/2023.emnlp-main.741},
  url = {https://aclanthology.org/2023.emnlp-main.741/}
}

@inproceedings{wei-etal-2024-longform,
  title = {Long-form Factuality in Large Language Models},
  author = {Wei, Jerry and Yang, Chengrun and Song, Xinying and Lu, Yifeng and Hu, Nathan and Huang, Jie and Tran, Dustin and Peng, Daiyi and Liu, Ruibo and Huang, Da and Du, Cosmo and Le, Quoc V.},
  booktitle = {Advances in Neural Information Processing Systems 37},
  year = {2024},
  doi = {10.52202/079017-2567},
  url = {https://proceedings.neurips.cc/paper_files/paper/2024/hash/937ae0e83eb08d2cb8627fe1def8c751-Abstract-Conference.html}
}

@inproceedings{cohan-etal-2019-structural,
  title = {Structural Scaffolds for Citation Intent Classification in Scientific Publications},
  author = {Cohan, Arman and Ammar, Waleed and van Zuylen, Madeleine and Cady, Field},
  booktitle = {Proceedings of the 2019 Conference of the North American Chapter of the Association for Computational Linguistics: Human Language Technologies, Volume 1 (Long and Short Papers)},
  pages = {3586--3596},
  address = {Minneapolis, Minnesota},
  publisher = {Association for Computational Linguistics},
  year = {2019},
  doi = {10.18653/v1/N19-1361},
  url = {https://aclanthology.org/N19-1361/}
}

@inproceedings{bowman-etal-2015-large,
  title = {A Large Annotated Corpus for Learning Natural Language Inference},
  author = {Bowman, Samuel R. and Angeli, Gabor and Potts, Christopher and Manning, Christopher D.},
  booktitle = {Proceedings of the 2015 Conference on Empirical Methods in Natural Language Processing},
  pages = {632--642},
  address = {Lisbon, Portugal},
  publisher = {Association for Computational Linguistics},
  year = {2015},
  doi = {10.18653/v1/D15-1075},
  url = {https://aclanthology.org/D15-1075/}
}

@inproceedings{thorne-etal-2018-fever,
  title = {{FEVER}: A Large-scale Dataset for Fact Extraction and {VER}ification},
  author = {Thorne, James and Vlachos, Andreas and Christodoulopoulos, Christos and Mittal, Arpit},
  booktitle = {Proceedings of the 2018 Conference of the North American Chapter of the Association for Computational Linguistics: Human Language Technologies, Volume 1 (Long Papers)},
  pages = {809--819},
  address = {New Orleans, Louisiana},
  publisher = {Association for Computational Linguistics},
  year = {2018},
  doi = {10.18653/v1/N18-1074},
  url = {https://aclanthology.org/N18-1074/}
}

@inproceedings{yona-etal-2024-large,
  title = {Can Large Language Models Faithfully Express Their Intrinsic Uncertainty in Words?},
  author = {Yona, Gal and Aharoni, Roee and Geva, Mor},
  booktitle = {Proceedings of the 2024 Conference on Empirical Methods in Natural Language Processing},
  pages = {7752--7764},
  address = {Miami, Florida, USA},
  publisher = {Association for Computational Linguistics},
  year = {2024},
  doi = {10.18653/v1/2024.emnlp-main.443},
  url = {https://aclanthology.org/2024.emnlp-main.443/}
}

@inproceedings{attributionbench2024,
  title = {{A}ttribution{B}ench: How Hard is Automatic Attribution Evaluation?},
  author = {Li, Yifei and Yue, Xiang and Liao, Zeyi and Sun, Huan},
  booktitle = {Findings of the Association for Computational Linguistics: ACL 2024},
  pages = {14919--14935},
  address = {Bangkok, Thailand},
  publisher = {Association for Computational Linguistics},
  month = aug,
  year = {2024},
  doi = {10.18653/v1/2024.findings-acl.886},
  url = {https://aclanthology.org/2024.findings-acl.886/}
}

@inproceedings{malaviya-etal-2024-expertqa,
  title = {{E}xpert{QA}: Expert-Curated Questions and Attributed Answers},
  author = {Malaviya, Chaitanya and Lee, Subin and Chen, Sihao and Sieber, Elizabeth and Yatskar, Mark and Roth, Dan},
  booktitle = {Proceedings of the 2024 Conference of the North American Chapter of the Association for Computational Linguistics: Human Language Technologies (Volume 1: Long Papers)},
  pages = {3025--3045},
  address = {Mexico City, Mexico},
  publisher = {Association for Computational Linguistics},
  month = jun,
  year = {2024},
  doi = {10.18653/v1/2024.naacl-long.167},
  url = {https://aclanthology.org/2024.naacl-long.167/}
}

@inproceedings{chen-etal-2024-understanding,
  title = {Understanding Retrieval Augmentation for Long-Form Question Answering},
  author = {Chen, Hung-Ting and Xu, Fangyuan and Arora, Shane and Choi, Eunsol},
  booktitle = {Proceedings of the 1st Conference on Language Modeling},
  year = {2024},
  url = {https://openreview.net/forum?id=j3AAkO5xgr},
  note = {{COLM} 2024}
}

@inproceedings{liu-etal-2023-evaluating,
  title = {Evaluating Verifiability in Generative Search Engines},
  author = {Liu, Nelson and Zhang, Tianyi and Liang, Percy},
  booktitle = {Findings of the Association for Computational Linguistics: EMNLP 2023},
  pages = {7001--7025},
  address = {Singapore},
  publisher = {Association for Computational Linguistics},
  month = dec,
  year = {2023},
  doi = {10.18653/v1/2023.findings-emnlp.467},
  url = {https://aclanthology.org/2023.findings-emnlp.467/}
}

@misc{bohnet-etal-2023-attributedqa,
  title = {Attributed Question Answering: Evaluation and Modeling for Attributed Large Language Models},
  author = {Bohnet, Bernd and Tran, Vinh Q. and Verga, Pat and Aharoni, Roee and Andor, Daniel and Baldini Soares, Livio and Ciaramita, Massimiliano and Eisenstein, Jacob and Ganchev, Kuzman and Herzig, Jonathan and Hui, Kai and Kwiatkowski, Tom and Ma, Ji and Ni, Jianmo and Sestorain Saralegui, Lierni and Schuster, Tal and Cohen, William W. and Collins, Michael and Das, Dipanjan and Metzler, Donald and Petrov, Slav and Webster, Kellie},
  year = {2023},
  eprint = {2212.08037},
  archivePrefix = {arXiv},
  primaryClass = {cs.CL},
  doi = {10.48550/arXiv.2212.08037},
  url = {https://arxiv.org/abs/2212.08037}
}

@inproceedings{sorodoc-etal-2025-garage,
  title = {{G}a{RAG}e: A Benchmark with Grounding Annotations for {RAG} Evaluation},
  author = {Sorodoc, Ionut Teodor and Ribeiro, Leonardo F. R. and Blloshmi, Rexhina and Davis, Christopher and de Gispert, Adri{\`a}},
  booktitle = {Findings of the Association for Computational Linguistics: ACL 2025},
  pages = {17030--17049},
  address = {Vienna, Austria},
  publisher = {Association for Computational Linguistics},
  month = jul,
  year = {2025},
  doi = {10.18653/v1/2025.findings-acl.875},
  url = {https://aclanthology.org/2025.findings-acl.875/},
  isbn = {979-8-89176-256-5}
}

@inproceedings{ragchecker2024,
  title = {{RAGChecker}: A Fine-Grained Framework for Diagnosing Retrieval-Augmented Generation},
  author = {Ru, Dongyu and Qiu, Lin and Hu, Xiangkun and Zhang, Tianhang and Shi, Peng and Chang, Shuaichen and {Cheng Jiayang} and Wang, Cunxiang and Sun, Shichao and Li, Huanyu and Zhang, Zizhao and Wang, Binjie and Jiang, Jiarong and He, Tong and Wang, Zhiguo and Liu, Pengfei and Zhang, Yue and Zhang, Zheng},
  booktitle = {Advances in Neural Information Processing Systems 37},
  year = {2024},
  doi = {10.52202/079017-0692},
  url = {https://proceedings.neurips.cc/paper_files/paper/2024/hash/27245589131d17368cccdfa990cbf16e-Abstract-Datasets_and_Benchmarks_Track.html}
}

@inproceedings{citeeval2025,
  title = {{C}ite{E}val: Principle-Driven Citation Evaluation for Source Attribution},
  author = {Xu, Yumo and Qi, Peng and Chen, Jifan and Liu, Kunlun and Han, Rujun and Liu, Lan and Min, Bonan and Castelli, Vittorio and Gupta, Arshit and Wang, Zhiguo},
  booktitle = {Proceedings of the 63rd Annual Meeting of the Association for Computational Linguistics (Volume 1: Long Papers)},
  pages = {32759--32778},
  address = {Vienna, Austria},
  publisher = {Association for Computational Linguistics},
  year = {2025},
  doi = {10.18653/v1/2025.acl-long.1574},
  url = {https://aclanthology.org/2025.acl-long.1574/}
}

@inproceedings{farkas-etal-2010-conll,
  title = {The {C}o{NLL}-2010 Shared Task: Learning to Detect Hedges and their Scope in Natural Language Text},
  author = {Farkas, Rich{\'a}rd and Vincze, Veronika and M{\'o}ra, Gy{\"o}rgy and Csirik, J{\'a}nos and Szarvas, Gy{\"o}rgy},
  booktitle = {Proceedings of the Fourteenth Conference on Computational Natural Language Learning -- Shared Task},
  pages = {1--12},
  address = {Uppsala, Sweden},
  publisher = {Association for Computational Linguistics},
  year = {2010},
  url = {https://aclanthology.org/W10-3001/}
}

@article{sauri-pustejovsky-2009-factbank,
  title = {{FactBank}: A Corpus Annotated with Event Factuality},
  author = {Saur{\'i}, Roser and Pustejovsky, James},
  journal = {Language Resources and Evaluation},
  volume = {43},
  number = {3},
  pages = {227--268},
  year = {2009},
  doi = {10.1007/s10579-009-9089-9},
  url = {https://doi.org/10.1007/s10579-009-9089-9}
}

@inproceedings{pustejovsky-etal-2003-timeml,
  title = {{TimeML}: Robust Specification of Event and Temporal Expressions in Text},
  author = {Pustejovsky, James and Casta{\~n}o, Jos{\'e} and Ingria, Robert and Saur{\'i}, Roser and Gaizauskas, Robert and Setzer, Andrea and Katz, Graham and Radev, Dragomir},
  booktitle = {New Directions in Question Answering: Papers from the 2003 {AAAI} Spring Symposium},
  publisher = {{AAAI} Press},
  year = {2003},
  url = {https://aaai.org/papers/0005-ss03-07-005-timeml-robust-specification-of-event-and-temporal-expressions-in-text/}
}

@article{sumner-etal-2014-exaggeration,
  title = {The Association between Exaggeration in Health Related Science News and Academic Press Releases: Retrospective Observational Study},
  author = {Sumner, Petroc and Vivian-Griffiths, Solveiga and Boivin, Jacky and Williams, Andy and Venetis, Christos A. and Davies, Aim{\'e}e and Ogden, Jack and Whelan, Leanne and Hughes, Bethan and Dalton, Bethan and Boy, Fred and Chambers, Christopher D.},
  journal = {BMJ},
  volume = {349},
  pages = {g7015},
  year = {2014},
  doi = {10.1136/bmj.g7015},
  url = {https://www.bmj.com/content/349/bmj.g7015}
}

@inproceedings{aly-etal-2021-feverous,
  title = {{FEVEROUS}: Fact Extraction and Verification Over Unstructured and Structured Information},
  author = {Aly, Rami and Guo, Zhijiang and Schlichtkrull, Michael and Thorne, James and Vlachos, Andreas and Christodoulopoulos, Christos and Cocarascu, Oana and Mittal, Arpit},
  booktitle = {Advances in Neural Information Processing Systems Datasets and Benchmarks Track},
  year = {2021},
  url = {https://arxiv.org/abs/2106.05707}
}

@misc{sharma2024domainqa,
  title = {Retrieval Augmented Generation for Domain-Specific Question Answering},
  author = {Sharma, Sanat and Yoon, David Seunghyun and Dernoncourt, Franck and Sultania, Dewang and Bagga, Karishma and Zhang, Mengjiao and Bui, Trung and Kotte, Varun},
  year = {2024},
  eprint = {2404.14760},
  archivePrefix = {arXiv},
  primaryClass = {cs.CL},
  doi = {10.48550/arXiv.2404.14760},
  url = {https://arxiv.org/abs/2404.14760}
}

@misc{cheng2026resolvingrobustnessprecisiontradeofffinancial,
  title = {Resolving the Robustness-Precision Trade-off in Financial {RAG} through Hybrid Document-Routed Retrieval},
  author = {Cheng, Zhiyuan and Lai, Longying and Liu, Yue},
  year = {2026},
  eprint = {2603.26815},
  archivePrefix = {arXiv},
  primaryClass = {cs.CL},
  url = {https://arxiv.org/abs/2603.26815}
}

@misc{cheng2026enhancingfinancialreportquestionanswering,
  title = {Enhancing Financial Report Question-Answering: A Retrieval-Augmented Generation System with Reranking Analysis},
  author = {Cheng, Zhiyuan and Lai, Longying and Liu, Yue and Cheng, Kai and Qi, Xiaoxi},
  year = {2026},
  eprint = {2603.16877},
  archivePrefix = {arXiv},
  primaryClass = {cs.CL},
  url = {https://arxiv.org/abs/2603.16877}
}

@article{zang2025reward,
  title = {Reward Auditor: Inference on Reward Modeling Suitability in Real-World Perturbed Scenarios},
  author = {Zang, Jianxiang and Wei, Yongda and Bai, Ruxue and Jiang, Shiyu and Mo, Nijia and Li, Binhong and Sun, Qiang and Liu, Hui},
  journal = {arXiv preprint arXiv:2512.00920},
  year = {2025},
  url = {https://arxiv.org/abs/2512.00920}
}

@article{zang2025alleviating,
  title = {Alleviating Attention Hacking in Discriminative Reward Modeling through Interaction Distillation},
  author = {Zang, Jianxiang},
  journal = {arXiv preprint arXiv:2508.02618},
  year = {2025},
  url = {https://arxiv.org/abs/2508.02618}
}

@article{lin2026reflect,
  title = {{Reflect-Guard}: Enhancing {LLM} Safeguards against Adversarial Prompts via Logical Self-Reflection},
  author = {Lin, Lixing and You, Juli and Li, Yue and Lin, Luyun and Wang, Yiqing and Zhang, Zhen and Zheng, Moxuan},
  journal = {arXiv preprint arXiv:2605.24834},
  year = {2026},
  eprint = {2605.24834},
  archivePrefix = {arXiv},
  primaryClass = {cs.CL},
  doi = {10.48550/arXiv.2605.24834},
  url = {https://doi.org/10.48550/arXiv.2605.24834}
}

@article{cai2025tone,
  title = {Does Tone Change the Answer? Evaluating Prompt Politeness Effects on Modern {LLMs}: {GPT}, {Gemini}, and {LLaMA}},
  author = {Cai, Hanyu and Shen, Binqi and Jin, Lier and Hu, Lan and Fan, Xiaojing},
  journal = {arXiv preprint arXiv:2512.12812},
  year = {2025},
  doi = {10.48550/arXiv.2512.12812},
  url = {https://arxiv.org/abs/2512.12812}
}

@inproceedings{li2025catch,
  title = {{CATCH}: A Modular Cross-Domain Adaptive Template with Hook},
  author = {Li, Xinjin and Lu, Yulie and Cao, Jinghan and Ma, Yu and Li, Zhenglin and Zhou, Yeyang},
  booktitle = {Advances in Visual Computing: 20th International Symposium, {ISVC} 2025, Las Vegas, NV, USA, November 17--19, 2025, Proceedings, Part I},
  series = {Lecture Notes in Computer Science},
  volume = {16396},
  pages = {41--52},
  publisher = {Springer},
  year = {2026},
  doi = {10.1007/978-3-032-14492-8_4},
  url = {https://doi.org/10.1007/978-3-032-14492-8_4}
}

\appendix

\section{Positioning Against Adjacent Evaluation Lines}
\label{sec:appendix-positioning}

\begingroup
\small
\paragraph{Attribution.}
Attribution work asks whether generated claims are supported by identified sources \citep{rashkin-etal-2023-measuring,yue-etal-2023-automatic}; \forcebench{} holds the source fixed and tests whether support decreases when only claim force becomes stronger.
\paragraph{RAG diagnostics.}
RAG diagnostics evaluate retrieval quality, answer correctness, and citation quality across full outputs \citep{gao-etal-2023-enabling,ragchecker2024,citeeval2025}; \forcebench{} isolates a local failure that can survive relevant retrieval and plausible citation placement.
\paragraph{Source audits.}
Source-audit settings check whether a citation is accessible, topically relevant, or factually consistent with the cited content; \forcebench{} assumes the passage is available, then asks whether the attached claim is calibrated to that passage's warrant.
\paragraph{Overstatement.}
Scientific-overstatement work studies exaggeration or evidential proportionality relative to source evidence \citep{sumner-etal-2014-exaggeration}; \forcebench{} transfers that concern to generated cited answers and operationalizes it as local contrastive pairs under one fixed citation.
\par
\endgroup

\section{Dataset and Annotation Documentation}
\label{sec:appendix-data}

\begingroup
\small
\paragraph{Sampling and candidate drafting.}
Sampling is designed to build stress-test examples, with prevalence estimation outside scope. We sample cited claim-evidence contexts from existing cited QA and attribution resources and keep contexts only when the displayed passage contains a visible local boundary in relation, modality, scope, temporal status, or numeric specificity. Candidate drafting then normalizes each selected context into the same row format: question or answer context, cited passage, decisive evidence span, evidence-calibrated claim, force-raised claim, repair, rationale, and proposed axis/severity metadata. Pre-annotation drafts are excluded before they become candidate rows if the passage is unavailable, the proposed contrast requires outside knowledge, the claims change the entity or answer target, or the contrast is an ordinary unsupported addition rather than a local force increase. All candidate rows that enter the annotation pool are double-annotated.

\paragraph{Candidate accounting.}
The benchmark is the result of one candidate-drafting, double-annotation, and adjudication workflow under a single local-warrant protocol. The same annotation pass is used regardless of source or date: every candidate row is independently reviewed by two research assistants before adjudication. In total, 433 candidate rows enter the annotation pool and 283 are accepted after adjudication and quality control. We then cap surplus accepted rows from overrepresented axes before benchmark reporting, select the headline evaluation set by stratifying over force axis and source provenance, and apply a final locality filter. Table~\ref{tab:construction_flow} gives the compact accounting for the annotation pool and headline denominator.

\begin{center}
\scriptsize
\setlength{\tabcolsep}{3pt}
\begin{tabular*}{\columnwidth}{@{\extracolsep{\fill}}p{0.72\columnwidth}r@{}}
\toprule
Counted set & Rows \\
\midrule
Candidate rows entering double annotation & 433 \\
Accepted after adjudication and quality control & 283 \\
Headline evaluation set before final locality filter & 200 \\
Final headline evaluation set after locality filter & 198 \\
\bottomrule
\end{tabular*}
\captionof{table}{\textbf{Construction flow.}
All candidate rows use the same two-annotator workflow. The final row is the denominator for headline model and baseline results.}
\label{tab:construction_flow}
\end{center}

\paragraph{Human review statistics.}
Candidate suggestions become benchmark labels only after human review. All 433 candidate rows that enter the annotation pool are independently annotated by two research assistants and then adjudicated before release; 283 are accepted and 150 are rejected. On pre-adjudication row-level decision labels, the annotators achieve Cohen's $\kappa=0.78$.

Every row in the final benchmark is therefore backed by two independent local-warrant judgments and final adjudication. Disagreements over accept/reject status, evidence-span length, safe-claim wording, overclaim wording, repair, primary axis, or severity are resolved before export. We treat axis and severity as adjudicated metadata: the axis is used for analysis and splitting, while severity is released for diagnosis; \mvr{} and \fs{} use only pair ordering.
\par
\endgroup

\section{Human Annotation Guidelines}
\label{sec:appendix-annotation}

The annotation guideline given to the research assistants defines the task as local citation-warrant judgment. Annotators are told to use only the displayed cited passage and to ignore whether a claim might be true according to outside knowledge.

\begingroup
\small
\paragraph{Annotation unit.}
Each row contains a question or answer context, a cited passage, a proposed evidence-calibrated claim, a proposed force-raised claim, and candidate metadata. Annotators may correct the evidence span, calibrated claim, force-raised claim, repair, primary axis, and severity in the \texttt{annotation\_*} fields, or reject the row.

\paragraph{Annotation interface.}
Annotators see the question or answer context and the cited passage, which is the only evidence allowed. They also see pre-annotation suggestions for the evidence span, calibrated claim, force-raised claim, repair, rationale, axis, and severity. Annotators write the release-facing fields in \texttt{annotation\_*}: row decision (\texttt{accept\_force\_gap}, \texttt{reject\_not\_force\_gap}, or \texttt{needs\_discussion}), corrected evidence span, calibrated claim, force-raised claim, repair, primary axis, severity, boundary type, and notes. Final labels come from these human corrections and adjudication.

\begin{appendixbox}{Guideline: Decision Procedure}
Annotators follow the same sequence for every candidate:
\begin{enumerate}[leftmargin=*, itemsep=1pt, topsep=2pt]
    \item Read the cited passage and mark the shortest span that determines the support boundary.
    \item Decide whether the calibrated claim is directly warranted by that span using only the displayed passage.
    \item Decide whether the force-raised claim preserves the same entity, event, source, and answer context.
    \item Decide whether the force-raised claim exceeds the passage's warrant on one primary axis.
    \item Assign relation, modality, scope, temporal, or numeric as the primary axis; mark severity; and write a repair that restores citation-warranted wording.
\end{enumerate}
\end{appendixbox}

\begin{appendixbox}{Guideline: Accept Criteria}
A row is accepted only when the calibrated claim is supported, the force-raised claim is under-warranted by the cited passage, and the contrast is local: it keeps the same core proposition while increasing force along one dominant axis. The annotator must be able to explain the contrast in the form: the evidence licenses X, but the force-raised claim states stronger Y.
\end{appendixbox}

\begin{appendixbox}{Guideline: Reject Criteria}
Annotators reject rows when the calibrated claim is unsupported; the force-raised claim is actually supported; the pair changes the main entity, event, time, or answer target; the row combines multiple force shifts with no clear primary axis; the stronger claim adds an unrelated unsupported fact; the judgment depends on outside knowledge; or the cited passage lacks a decisive evidence span.
\end{appendixbox}

\paragraph{Decision labels.}
Annotators choose one row-level decision. \texttt{accept\_force\_gap} means the citation supports the calibrated claim, the stronger claim exceeds that same citation, and the contrast is local. \texttt{reject\_not\_force\_gap} means the row is unsupported, already supported, nonlocal, wrong-entity\slash topic, or otherwise outside scope. \texttt{needs\_discussion} is reserved for genuinely unclear boundaries and is resolved during adjudication before release.

\paragraph{Axis definitions.}
Relation errors turn association, mention, risk, or correlation into causation, proof, prevention, or obligation. Modality errors turn possibility, uncertainty, suggestion, or conditional evidence into certainty or necessity. Scope errors generalize from a bounded population, product, jurisdiction, study, source, or version to a broader class. Temporal errors turn dated, future, predicted, or as-of evidence into current or timeless claims. Numeric errors turn approximate, ranged, bounded, or comparative quantities into exact point claims.

\paragraph{Severity and repair.}
Severity 1 marks a subtle wording shift that can still mislead an evaluator; severity 2 marks a clear force mismatch; severity 3 marks a high-impact or strongly misleading mismatch. Repairs must preserve only what the cited passage warrants, usually by restoring a qualifier, scope restriction, time marker, approximate quantity, or weaker relation.
\par
\endgroup

\section{Representative Benchmark Examples}
\label{sec:appendix-examples}

Each example below shows one retained benchmark row after the final locality filter. The cited evidence is fixed; the evidence-calibrated claim is licensed by that evidence, while the force-raised claim asks the citation to support stronger force than it states.

\begingroup
\small
\paragraph{Relation.}
Evidence cue: ``associated with up to a 40 per cent risk.''
Calibrated claim: The use of sodium valproate during pregnancy is associated with up to a 40 percent risk of neurodevelopmental disorders and a 10 percent risk of physical disabilities for an unborn child.
Force-raised claim: The use of sodium valproate during pregnancy results in up to a 40 percent risk of neurodevelopmental disorders and a 10 percent risk of physical disabilities for an unborn child.

\paragraph{Modality.}
Evidence cue: ``evidence that men may have been more susceptible.''
Calibrated claim: Some evidence suggests men may have been more susceptible to the 1361--1362 plague.
Force-raised claim: Men were more susceptible to the 1361--1362 plague.

\paragraph{Scope.}
Evidence cue: ``convulsive status epilepticus in children.''
Calibrated claim: A trial found levetiracetam was not superior to phenytoin as a second-line agent for convulsive status epilepticus in children.
Force-raised claim: A trial found levetiracetam was not superior to phenytoin as a second-line agent for convulsive status epilepticus patients.

\paragraph{Temporal.}
Evidence cue: ``As of 2019, New Jersey has nine casinos.''
Calibrated claim: As of 2019, New Jersey had nine casinos, all in Atlantic City.
Force-raised claim: There are currently nine casinos in Atlantic City, New Jersey.

\paragraph{Numeric.}
Evidence cue: ``typically sits around 1.5 to 2.0.''
Calibrated claim: Ebola's R0 typically sits around 1.5 to 2.0.
Force-raised claim: Ebola's R0 is 2.0.
\par
\endgroup

\section{Boundary Accounting}
\label{sec:appendix-boundary}

The main text treats rejected rows as boundary evidence that explains exclusions. These rows are important because they show that the guideline separates local force gaps from ordinary unsupported claims, supported stronger claims, and candidate-construction artifacts.

\begingroup
\small
\paragraph{Candidate boundary checks.}
Candidate rows are screened for local admissibility before release. Rejection categories include unsupported calibrated claims, supported force-raised claims, reversed or nonlocal relation rewrites, duplicate rows, safe or underclaimed rows, ordinary unsupported additions, ambiguous force gaps, and evidence spans that are not decisive.
\paragraph{Independent annotation checks.}
All candidate rows that enter the annotation pool are independently double-annotated before adjudication; pre-adjudication row-level agreement is Cohen's $\kappa=0.78$. Accept/reject disagreements are resolved before inclusion. Axis checks are computed only after both annotators accept a row, so they are treated as conditional consistency checks over accepted rows.
\par
\endgroup

These checks document the boundary between local citation-force gaps and other citation or candidate-construction failures; prevalence estimation remains outside scope.

\section{Judge Prompt Templates}
\label{sec:appendix-prompts}

The exact machine-readable prompt files are released with the benchmark. The templates below show the prompts used for the reported model-judge results.

\begingroup
\small
\begin{appendixbox}{Generic Support Judge}
\textbf{System.} ``You are evaluating whether a cited passage supports a claim. Use only the cited passage. Do not use outside knowledge.''

\textbf{User.} question, cited passage, and claim, followed by: ``Does the cited passage support the claim? Choose exactly one label: \texttt{fully\_supported}, \texttt{partially\_supported}, \texttt{unsupported}, \texttt{contradicted}, \texttt{irrelevant}. Return JSON only with fields \texttt{label}, \texttt{confidence}, and \texttt{rationale}.''
\end{appendixbox}

\begin{appendixbox}{Force-Aware Judge}
\textbf{System.} ``You are evaluating evidence-force calibration. Use only the cited passage. Do not use outside knowledge. A claim is force-calibrated only if the cited evidence warrants the claim at the same or stronger level of relation, modality, scope, temporal validity, and numeric specificity.''

\textbf{User.} question, cited passage, and claim, followed by: ``Identify the strongest claim warranted by the passage; decide whether the given claim exceeds the passage's warranted force; if it exceeds the evidence, identify the primary force-gap axis. Choose exactly one label: \texttt{force\_calibrated}, \texttt{force\_gap}, \texttt{contradicted}, \texttt{irrelevant}. If \texttt{force\_gap}, choose one primary axis: \texttt{relation}, \texttt{modality}, \texttt{scope}, \texttt{temporal}, \texttt{numeric}. Return JSON only with fields \texttt{label}, \texttt{primary\_axis}, \texttt{severity}, \texttt{warranted\_claim}, and \texttt{rationale}.''
\end{appendixbox}

\begin{appendixbox}{Axis-Free Force Ablation}
\textbf{System.} ``You are evaluating whether a cited passage warrants a claim as written. Use only the cited passage. Do not use outside knowledge.''

\textbf{User.} question, cited passage, and claim, followed by: ``Decide whether the cited passage licenses the claim exactly as worded. Mark \texttt{force\_gap} if the claim is related to the passage but says more than the passage licenses. Mark \texttt{force\_calibrated} only if the cited passage licenses the full strength of the claim as written. Do not use a predefined taxonomy. Return JSON only with fields \texttt{label}, \texttt{primary\_axis}, \texttt{severity}, \texttt{warranted\_claim}, and \texttt{rationale}.''
\end{appendixbox}

\begin{appendixbox}{Dummy-Axis Rubric Control}
\textbf{System.} ``You are a citation judge. Use only the cited passage. Do not use outside knowledge.''

\textbf{User.} question, cited passage, and claim, followed by: ``Decide whether the cited passage supports the claim. When checking support, focus on source authority, writing clarity, citation placement, topical completeness, and fluency. Choose exactly one label: \texttt{force\_calibrated}, \texttt{force\_gap}, \texttt{contradicted}, \texttt{irrelevant}. Return JSON only with fields \texttt{label}, \texttt{primary\_axis}, \texttt{severity}, \texttt{warranted\_claim}, and \texttt{rationale}.''
\end{appendixbox}
\par
\endgroup

\section{Model API and Decoding Configuration}
\label{app:model_config}

All reported model API calls were executed on May 17, 2026 UTC. All four model judges were called through their API endpoints. The exact model identifiers were \texttt{gpt-5.5}, \texttt{claude-sonnet-4-6}, \texttt{glm-5.1}, and \texttt{qwen3.6-flash}.

For reproducibility, every claim judgment was submitted as a fresh stateless API request with deterministic decoding, JSON-only output instructions, and a 1024-token output cap. GPT-5.5 used the Responses-style \texttt{max\_output\_tokens=1024} setting; the other API calls used \texttt{max\_tokens=1024}. For chat-style APIs, deterministic decoding used \texttt{temperature=0.0}; for provider interfaces with a named deterministic mode, we used that mode. The same prompt templates, parser, scoring script, retry policy, and failure-handling rules were used for all models and prompts. Run manifests record the API target, exact model identifier, prompt file, request count, output cap, decoding settings, run timestamp, parser version, retry counts, and sanitized output paths.

\end{document}